\documentclass[letterpaper,10pt,conference]{ieeeconf}
\IEEEoverridecommandlockouts 
\usepackage{cite}
\usepackage{amsmath,amssymb,amsfonts}
\usepackage{algorithmic}
\usepackage{graphicx}
\usepackage{textcomp}
\usepackage{xcolor}
\usepackage{comment}
\usepackage{colortbl}
\usepackage{todonotes}
\usepackage{soul}
\usepackage[bookmarks=true]{hyperref}

\def\BibTeX{{\rm B\kern-.05em{\sc i\kern-.025em b}\kern-.08em
    T\kern-.1667em\lower.7ex\hbox{E}\kern-.125emX}}

\title{\LARGE \bf Online Localisation and Colored Mesh Reconstruction Architecture \\ for 3D Visual Feedback in Robotic Exploration Missions
}

\author{Quentin Serdel, Christophe Grand, Julien Marzat and Julien Moras 
\thanks{Q. Serdel, J. Marzat and J. Moras are with DTIS, ONERA, Université Paris-Saclay, 91123 Palaiseau, France. \newline
C. Grand is with DTIS, ONERA, University of Toulouse, France. \newline
Emails: \texttt{\{firstname.lastname\}@onera.fr}}}

\begin{document}
\maketitle

\begin{abstract}

This paper introduces an Online Localisation and Colored Mesh Reconstruction (OLCMR) ROS perception architecture for ground exploration robots aiming to perform robust Simultaneous Localisation And Mapping (SLAM) in challenging unknown environments and provide an associated colored 3D mesh representation in real time. It is intended to be used by a remote human operator to easily visualise the mapped environment during or after the mission or as a development base for further researches in the field of exploration robotics. The architecture is mainly composed of carefully-selected open-source ROS implementations of a LiDAR-based SLAM algorithm alongside a colored surface reconstruction procedure using a point cloud and RGB camera images projected into the 3D space. The overall performances are evaluated on the \textit{Newer College} handheld LiDAR-Vision reference dataset and on two experimental trajectories gathered on board of representative wheeled robots in respectively urban and countryside outdoor environments. 
\end{abstract}

\begin{keywords}
Field Robots, Mapping, SLAM, Colored Surface Reconstruction 
\end{keywords}

\section{Introduction}

Embedded architectures for autonomous robots performing exploration missions are continuously improving, such that localisation and map construction in a previously uncharted environment becomes possible with limited human intervention~\cite{lluvia2021active}. Many SLAM-based localisation methods have been proposed and extensively evaluated on reference datasets. They rely either on monocular cameras or stereovision~\cite{fuentes2015visual,yasuda2020autonomous,merzlyakov2021comparison}, or on 2D or 3D LiDAR scanners~\cite{elhousni2020survey,zou2021comparative,rtab-map,NC_3D_reconstruction}, with some recent attempts to combine both types of sensors~\cite{debeunne2020review}. Several computationally-efficient online mapping and 3D reconstruction approaches have been proposed in parallel~\cite{kim2018slam}, while the specific issue of the colourisation of mesh or point cloud representations has been investigated independently~\cite{waechter2014let,chan2021post}. However, there remains the need to further evaluate in realistic conditions the behaviour and performances of full systems which are able to combine online localisation, mapping and mesh colourisation on real ground robots equipped with 3D LiDAR and vision sensors~\cite{aguiar2020localization}. 

Early work on the subject of real-time 3D mesh reconstruction in \cite{early_online_mesh_construction} introduced real-time 3D surface mesh reconstruction in an urban environment using stereo camera images and pose estimation obtained by fusing GNSS measurements and visual odometry. The method proposed in \cite{chisel} performs real-time 3D mapping of house-sized indoor environments through the application of Truncated Signed Distance Function \cite{SDF} and dynamic voxel hashing \cite{hashing}, with Visual-Inertial Odometry (VIO) as localisation source. While 3D mesh reconstruction was mainly intended to be used as a visualisation tool for human operators, later work has focused on the aspect of surfacic mesh mapping for navigation purposes, arguing that surface-based maps contain more dense information compared to sparse point clouds. This makes them more exploitable for autonomous driving or robot navigation (either autonomous or remotely-operated). Limitation of memory usage and computational cost along with scalability of the 3D mesh reconstruction solution are major concerns when addressing the deployment of such systems on the field. 
In \cite{real_time_3D_mesh_reconstruction}, a method has been presented to carry out online 3D large-scale mesh reconstruction using manifold mapping and monocular-camera-based localisation applied to urban mapping and evaluated on the KITTI autonomous driving dataset~\cite{kitti}. An online localisation and dense scalable 3D map reconstruction architecture has been presented in~\cite{surfel_mapping} with grayscale coloring. It implements surfel based methods with the use of RGB-D, stereo and monocular cameras.
Texture projection on reconstructed meshes is addressed in \cite{urban-mapping} and \cite{mesh-reconstruction}, however the presented methods consist of a post-processing of the whole data and are thus performed offline.
In \cite{onera_multi_robot_mapping}, a multi-robot system using stereovision-based localisation and 3D TSDF manifold mapping with grayscale coloring has been shown to run in real-time in an indoor environment. 
A real-time approach for creating and maintaining a colourised or textured surface reconstruction from RGB-D sensors has been introduced in~\cite{ScalableFusion}, with a special focus on memory management for scalability.

\begin{figure}[!t]
\centerline{
\includegraphics[width=0.49\textwidth]{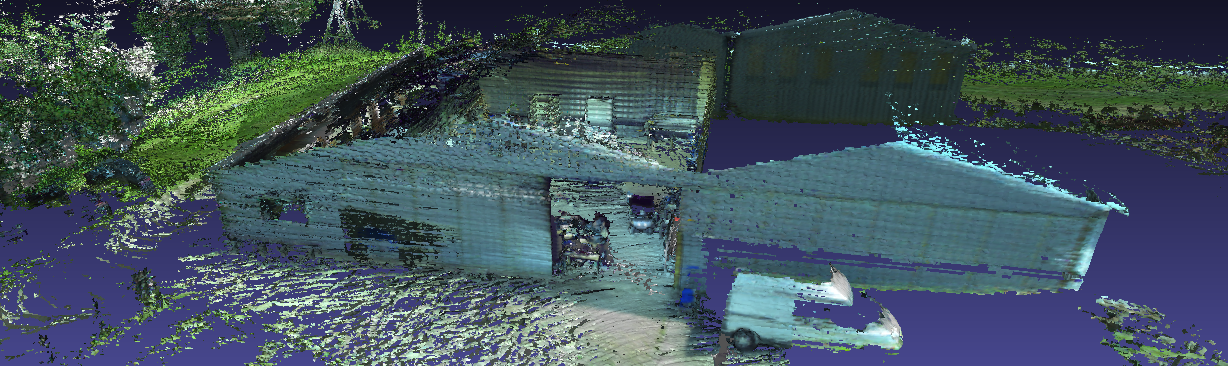}}
\vspace*{-3mm}
\caption{Example of online colored mesh rendered by the proposed system\label{final_render}}
\vspace*{-7mm}
\end{figure}

The Online Localisation and Colored Mesh Reconstruction (OLCMR) architecture proposed in the present paper is a complete system that performs both localisation and colored mesh reconstruction in real-time on board of a ground robot equipped with a 3D LiDAR scanner and multiple cameras. Figure~\ref{final_render} depicts an example of reconstruction produced by this system. The designed system builds upon recent open-source implementations of LiDAR-based SLAM and 3D mesh reconstruction methods that are summarized in Section~\ref{archi} in perspective with related work, along with a description of the adaptations that were necessary to tackle the common objective pursued here.
The overall performances of the proposed OLCMR architecture have then been evaluated on the handheld Newer College reference benchmark~\cite{newer_college,newer_college_2} and are reported in Section~\ref{NewerCollege}. The results and computational needs achieved on two experimental datasets acquired with tele-operated ground robots in urban and countryside outdoor environments are given in Section~\ref{OneraDatasets} to demonstrate the versatility and wide applicability of the system. The results on these three different test cases include the evaluation of localisation (with and without loop-closure) and mapping accuracy with respect to independent reference models. Illustrations of mesh coloration are also presented for each dataset alongside images taken from the robot camera to highlight the quality of the whole reconstruction.

\section{OLCMR Architecture description}\label{archi}

\begin{figure}[t!]
\centerline{\includegraphics[width=0.48\textwidth]{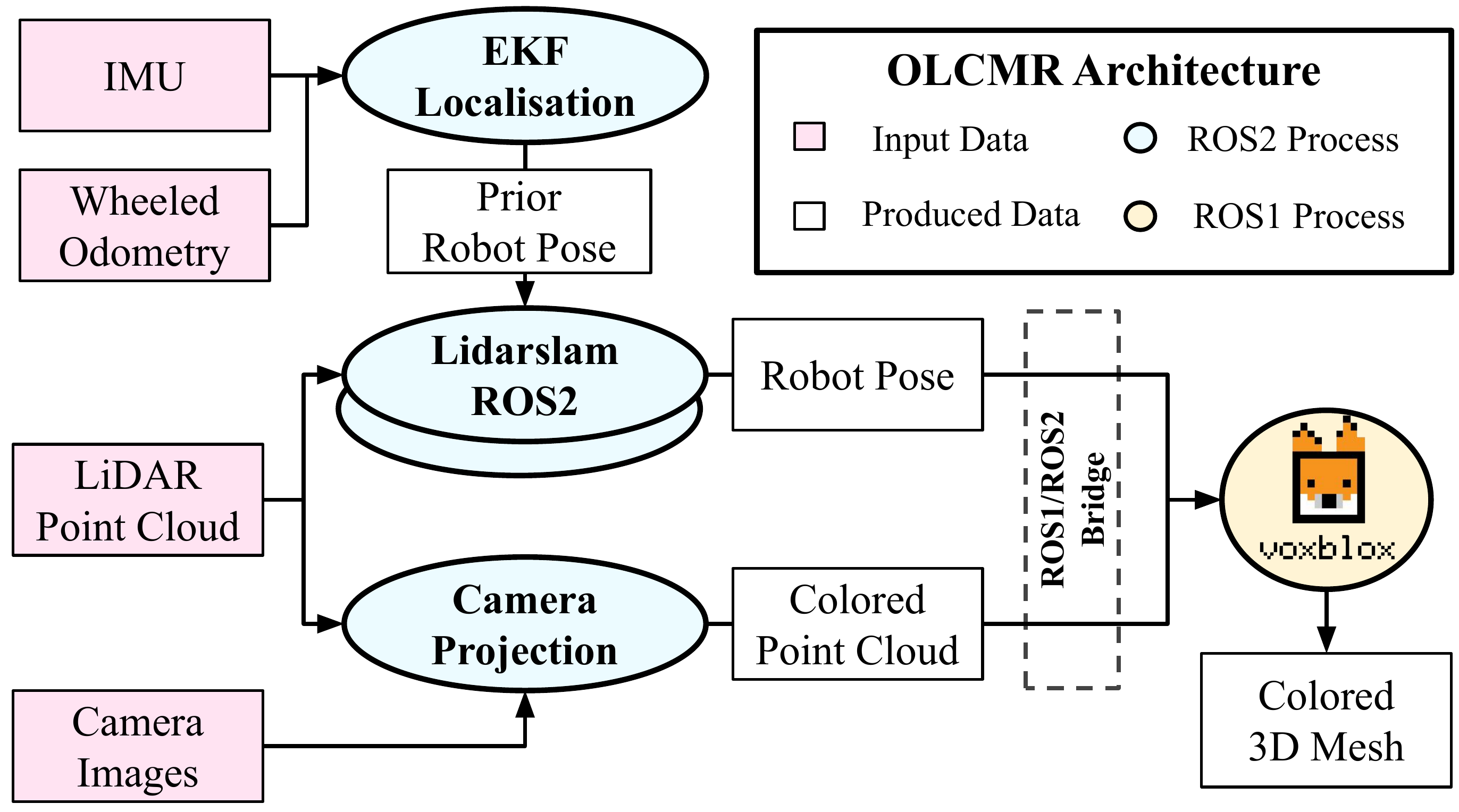}}
\caption{OLCMR architecture overview}
\vspace*{-5mm}
\label{OLTMR_arch}
\end{figure}

The system architecture (Figure~\ref{OLTMR_arch}) has been designed to process data on-board of a ground robot for online missions in diverse uncharted environments. The main requirement was to be able to combine data from a 3D Laser scanner, one or several monocular cameras and an IMU, given intrinsic and extrinsic calibration parameters of these sensors (using e.g.~\cite{kalibr}). The main objective is to compute online a dense colored 3D reconstruction with sufficient localisation accuracy, in order to be provided during the autonomous robot mission.
The architecture is intended to function in various types of scenarios presenting challenging characteristics such as GNSS-denied, unstructured surroundings, dim or varying lights, uneven terrain. 

It has been chosen to rely on a LiDAR-based SLAM (\textit{Lidarslam\_ROS2})\cite{LidarSlam} for localisation to be light-invariant and to navigate in environments with possibly long ranges to points of interest, thus ruling out most of vision-based methods~\cite{merzlyakov2021comparison,zou2021comparative}. The choice of the LiDAR as the main localization and mapping sensor also readily provides a 3D point cloud without any additional computational cost for the embedded CPU. IMU measurements and kinematic odometry (based on wheel encoders) are additionally used to robustify the robot localisation through EKF sensor fusion. The camera images are intended to enrich the produced 3D mesh by incorporating color levels or any kind of visually extracted data (e.g., semantic classification in a future perspective) through their projection in the 3D dense reconstruction using their relationship with the LiDAR point cloud. The TSDF-based mapping method~\textit{Voxblox}~\cite{voxblox} is used to obtain this 3D dense reconstruction of the environment  

The processing required by OLCMR is entirely performed on CPU (see related evaluation results in Section~\ref{OneraDatasets}). The architecture is implemented under ROS2 Galactic, with the 3D mesh reconstruction running under ROS Noetic and communicating through a ROS1/ROS2 bridge~\cite{stavrinos2021ros2}.

\subsection{Localisation}\label{lidarslam_ekf}


Reviews of open-source ROS SLAM implementations based on the use of stereo cameras, depth cameras or LiDAR are proposed in \cite{rtab-map,merzlyakov2021comparison,yasuda2020autonomous,elhousni2020survey}. 
For the previously stated reasons, our choice for a SLAM implementation has been restricted to the LiDAR based approaches. Since OLCMR is supposed to operate in complex outdoor environments (possibly unstructured and with uneven terrain) and relies on a 3D point cloud for mapping purposes, 2D LiDAR based approaches have been excluded. 
Recent SLAM implementations are often composed of two distinct parts. The SLAM front-end allows real-time localisation of the robot relative to its close environment, relying on high-rate sensors, either proprioceptive or external. The back-end estimates and corrects the localisation drift induced by the front-end over time. It run at a lower rate and relies either on absolute measurements such as GNSS as in LIO-SAM~\cite{liosam}, or landmarks of known absolute position or on the recognition of previously visited areas (loop-closure). The vast majority of LiDAR-based SLAM front-end implementations are centered on the use of a scan-to-scan or scan-to-submap matching algorithm. The most used approach for scan-to-scan matching is Iterative Closest Point (ICP)~\cite{recent_icp}, which however presents significant drawbacks in the studied context where the LiDAR clouds are composed of a vast number of points, depending on the sensor angular resolution and beam number (typical numbers being 16, 32, 64, 128). Thus, the sole application of an iterative, exhaustive algorithm such as ICP for scan matching can be suboptimal 
requiring a consequent amount of calculation to converge and grant satisfying results. A widely used solution to that issue is the extraction of interest points for each LiDAR scan using geometric properties of the cloud's distribution, as in LEGO-LOAM \cite{legoloam}. The ICP matching is then performed only using these more relevant points thus requiring way less computational capacity. However, the interest point extraction is still a costly process and requires the point cloud to present a somehow organised distribution for them to be extracted efficiently.
As an alternative, a stochastic method based on normal approximation of the point cloud distribution named Normal Distribution Transform (NDT) was introduced in~\cite{ndt}. This makes the SLAM algorithm appropriate for evolution in unstructured environment as well as in geometrically rich ones, and requires limited computational power with no feature point extraction processing.

For these reasons, the \textit{Lidarslam\_ROS2} open-source SLAM implementation of a  LiDAR scan-to-scan matching has been integrated in the proposed architecture.
In order to reduce the impact of locations that could be difficult to map (e.g. corridors with repeating geometrical patterns or bare flat fields) and for the SLAM to converge faster, the scan matcher takes a prior transform and differentiates it from the latest prior to use as initial guess for the transform between two scans. The prior choice is left to the discretion of the user. In the current case, the prior must present a good trade-off between accuracy and computational cost. It has thus been chosen to estimate it by fusing the forward velocity from wheeled odometry and the orientation and angular velocities from the IMU using an Extended Kalman Filter. The chosen method 
requires a significant overlap between successive scans, which could make it unsuitable to fast moving vehicles such as autonomous cars.
In order to avoid some local failures in the scan-to-scan pose estimation (typically happening during fast rotations or very jittery motions), we proposed the following procedure: when the estimated transform from the last scan pose is greater than a threshold (set to 0.5 m) from the prior pose, this prior is used as final estimated transform and the associated scan is not registered into the map. The LidarSlam back-end performs loop-closure detection by comparing the current scan to stored key-frames using NDT and pose graph optimisation with the g$^{2}$o framework \cite{g2o}.


\subsection{3D Dense Reconstruction}\label{voxblox}

The idea of producing a dense representation of the explored environment from the LiDAR point-clouds instead of using the sparse map produced by the SLAM implementation is driven by two different needs. In the first place, a point-cloud map is not the most adequate for visualisation by a human operator as the structure of mapped objects remains ambiguous when looking at it, especially in cluttered environments. Secondly, a dense representation is much more adapted to navigation purposes as it allows the robot to infer the terrain traversability at any coordinates without requiring further interpolation and therefore navigate in the full exploration space.  
Various methods for 3D dense representation of the environment have been developed by the robotic and computer vision community. Since OLCMR is intended to function in real-time, offline methods such as Structure from Motion and Poisson surface reconstruction have been dismissed. Voxel-based visualisation produced by methods such as Octomap~\cite{octomap} is well-suited for inclusion in autonomous navigation loops but less in terms of visualisation. Online surfacic methods such as the previously mentioned TSDF~\cite{SDF} use the information of sensor position relative to points in the cloud, thus removing the ambiguity of surface orientation and allowing to identify free space between the sensor and the mapped points. 
For these reasons, the 3D mesh reconstruction relies on the open-source implementation of the \textit{Voxblox} ROS package~\cite{voxblox} to build incrementally a surfacic mesh from each LiDAR scan using the generated point-cloud and the current robot pose (given by the SLAM/EKF process described in Section~\ref{lidarslam_ekf}) before fusing it in the globally reconstructed mesh. 

\subsection{2D-3D color re-projection}\label{colorisation}
A dedicated process handles the colourisation of the LiDAR points with the corresponding pixel values from the RGB camera images, for further inclusion in a 3D colored mesh.
The projection of the 3D points into the camera images is computed geometrically~\cite{projection} as follows. 
For each camera, the coordinates of the 3D LiDAR points are expressed into the camera frame, with $R$ and $t$ respectively being the rotation matrix and translation vector between the LiDAR ($L$) and camera ($C$) frames.
\begin{equation}
    \begin{bmatrix}
           x_i & y_i &  z_i
         \end{bmatrix}_C^T = R . \begin{bmatrix}
           x_i & y_i & z_i
         \end{bmatrix}_L^T + t
\end{equation}
The coordinates are normalised by the point depth values.
\begin{equation}
    \begin{bmatrix}
           x'_i \\
           y'_i \\
         \end{bmatrix} = \begin{bmatrix}
           1/z_i & 0 \\
           0 & 1/z_i \\
         \end{bmatrix} . \begin{bmatrix}
           x_i \\
           y_i \\   
         \end{bmatrix}_C
\end{equation}
The coordinates are projected into the image plane using the intrinsic camera matrix, where the parameters $p_x$, $p_y$, $c_u$ and $c_v$ come from the camera calibration process.
\begin{equation}
    \begin{bmatrix}
           u \\
           v \\
           1
         \end{bmatrix} =  \begin{bmatrix}
           p_x & 0 & c_u \\
           0 & p_y & c_v \\
           0 & 0 & 1
         \end{bmatrix} . \begin{bmatrix}
           x'_i \\
           y'_i \\
           1
         \end{bmatrix}
\end{equation}
Let $V(u,v)$ be the value (e.g. RGB color) of the pixel of coordinates $u,v$. Because cameras are affected by distortion, undistortion is applied to compute the image $U$ using the distortion model and parameters given during calibration. 
If the resulting coordinates are inside the image bounds, the RGB value of the pixel of coordinates $[u,v]$ is allocated to the corresponding 3D LiDAR point. The color field is denoted as $C$.
\begin{equation}
    C(x_i,y_i,z_i) = U(u,v)
\end{equation}
Each surface reconstructed by the TSDF is then colored by the \textit{Voxblox} pipeline with a recursive average filtering of its vertex colors.

For the evaluations of Section~\ref{NewerCollege}, the full calibration parameters were available in the reference dataset. For the robot acquisitions from Section~\ref{OneraDatasets}, the calibration process was performed using the Kalibr ROS package~\cite{kalibr}. The intrinsic parameters and distortion model are estimated for each camera using a collection of images of an AprilTag grid of known dimensions. Kalibr exploits IMU measurements and the camera overlaps to estimate their extrinsic parameters. The transform between the LiDAR and the cameras was determined manually using the CAD model of the robots. Note that it is well known that combined LiDAR-vision systems can be very sensitive to the calibration between these two sensors for re-projection or SLAM purposes~\cite{debeunne2020review}.


\section{Evaluation on the Newer College dataset}\label{NewerCollege}


{Robot-held public datasets such as \cite{marulan_dataset} or \cite{husky_dataset} do not contain all the required sensors at once (e.g. LiDAR and cameras). Vehicle-based datasets such as KITTI \cite{kitti} are not adapted to our architecture, with large moving objects and reduced overlap between successive scans.}
The performances of the OLCMR architecture have thus been evaluated in terms of localisation precision and 3D reconstruction quality with the Newer College Dataset \cite{newer_college} and its 2021 extension \cite{newer_college_2}, and a qualitative assessment of the mesh colourisation is also presented. The Newer College dataset extension  provides LiDAR scans, IMU measurements and monocular images gathered from 4 cameras aboard a handheld device along various trajectories inside New College, Oxford. Ground-truth for the evaluation of SLAM and 3D reconstruction are provided using a tripod-mounted survey LiDAR and ICP registration, and many modern SLAM solutions have been evaluated on this dataset. For these reasons, this dataset has been deemed to be relevant for the evaluation of the OLCMR architecture performances. Although the architecture has been developed to function optimally aboard a ground robot, a few changes allowed it to be efficient while treating the data gathered from these handheld trajectories. The robot kinematic odometry used as prior has been replaced with a constant forward speed of 1.0 $m/s$ and the 128 beam LiDAR point cloud has been down-sampled 10 times for SLAM input to limit its CPU usage. This section presents these evaluation results with relevant comparison to state-of-the-art. All evaluations were performed on a Intel Xeon(R) W-2123 8 core 3.60GHz CPU with 16 GB of RAM.

\subsection{Localisation Evaluation}
\begin{figure}[h!]
\vspace{-0.5cm}
\centerline{
\includegraphics[width=0.49\textwidth]{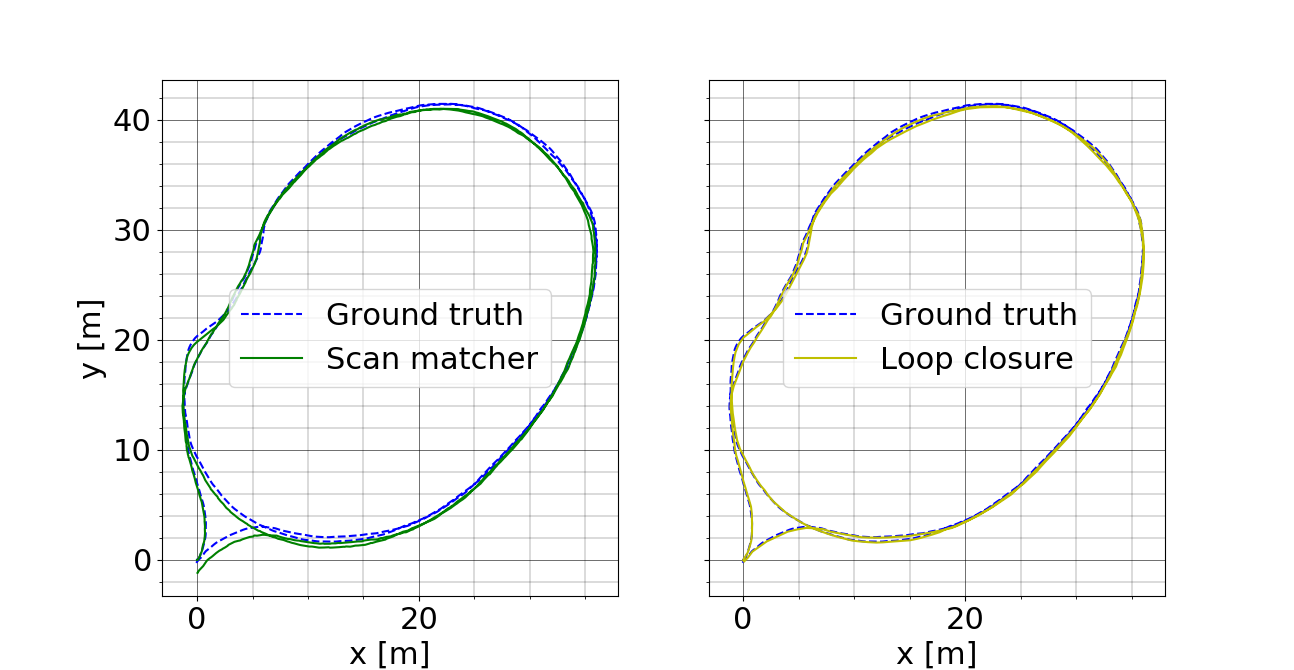}}
\centerline{
\includegraphics[width=0.49\textwidth]{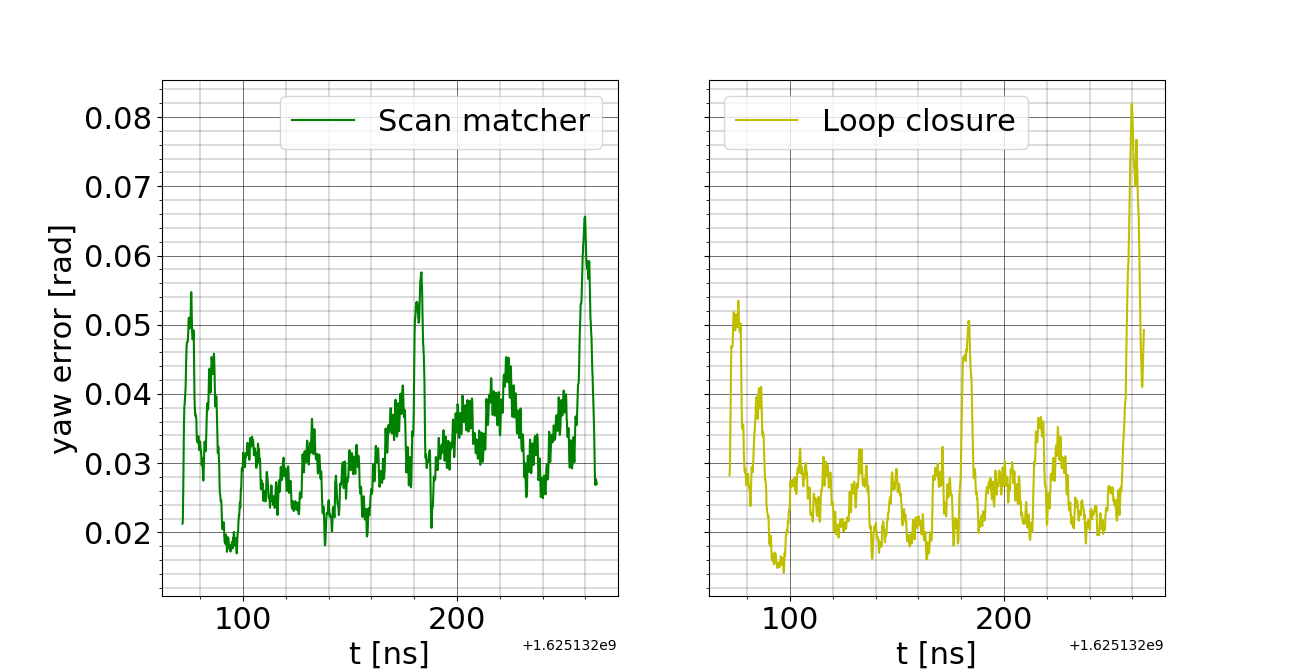}}
\caption{Newer College localisation evaluation, top shows trajectories estimated after scan matching (left) and loop closure optimisation (right) superposed to the ground truth trajectory. Bottom shows Absolute Yaw Error smoothed over 50 samples for same trajectories.}
\label{cloistre_slam_eval}
\vspace*{-5mm}
\end{figure}

The localisation building blocks (LiDAR SLAM and EKF management of prior) of the proposed architecture have been evaluated on this dataset. The goal of this evaluation is to ensure that this localisation is sufficiently precise to be used for the overall colored mesh reconstruction. Following the localisation evaluation protocol proposed in~\cite{slam_evaluation}, the Relative Pose Error over 10 m (RPE) and the Absolute Trajectory Error (ATE) are respectively computed for the \textit{2021 quad-easy} and \textit{2020 short-experiment} datasets. They are compared to state-of-the art SLAM evaluations on the same trajectories (as respectively reported in~\cite{nc_slam_compare} and \cite{quenzel2021realtime}). 
Fig. \ref{cloistre_slam_eval} shows the trajectories estimated by the SLAM scan matcher and loop closure optimiser superimposed on the ground truth trajectory, as well as yaw errors. Table \ref{cloistre_slam_compare_tab} summarizes the SLAM performances. 
The localisation performance is consistent with the best currently available SLAM methods and the loop closure yields a small performance improvement, since the front-end SLAM presented only a small drift in this richly-textured environment.

\begin{table}[htbp]
\caption{SLAM evaluation on the Newer College dataset}
\begin{center}
\begin{tabular}{|c|c|c|c|c|}
\hline
\multicolumn{2}{|c|}{\textbf{\textit{nc 2021 quad-easy}}} &\cellcolor{black}& \multicolumn{2}{c|}{\textbf{\textit{nc 2020 short-experiment}}}\\
\hline
\textbf{Duration} & 198.7 s &\cellcolor{black}& \textbf{Duration} & 1530.0 s \\
\hline
\textbf{Length} & 245.7 m &\cellcolor{black}& \textbf{Length} & 1570.3 m \\
\hline
\hline
\textbf{SLAM} & \textbf{RPE} &\cellcolor{black}& \textbf{SLAM} & \textbf{APE} \\
\hline
\hline
OpenVINS \cite{open_vins}& 1.01 m &\cellcolor{black}& A-LOAM \cite{LOAM}& 3.308 m \\
\hline
ORB-SLAM3 \cite{orb_slam3}& 0.23 m &\cellcolor{black}& F-LOAM \cite{F-LOAM}& 101.9 m \\
\hline 
VILENS-MC \cite{nc_slam_compare} & 0.31 m &\cellcolor{black}& MARS \cite{quenzel2021realtime}& 1.978 m \\
\hline 
VILENS-Stereo & 0.30 m &\cellcolor{black}& SuMa \cite{SuMa} & 2.048 m \\
\hline 
LidarSlam SM & 0.28 m &\cellcolor{black}& LidarSlam SM & 5.141 m \\
\hline
LidarSlam LC & 0.27 m &\cellcolor{black}& LidarSlam LC & 4.010 m \\
\hline
\end{tabular}
\label{cloistre_slam_compare_tab}
\end{center}
\end{table}

\subsection{3D Reconstruction Evaluation}
\begin{figure}[htbp]
\centerline{
\includegraphics[width=0.49\textwidth]{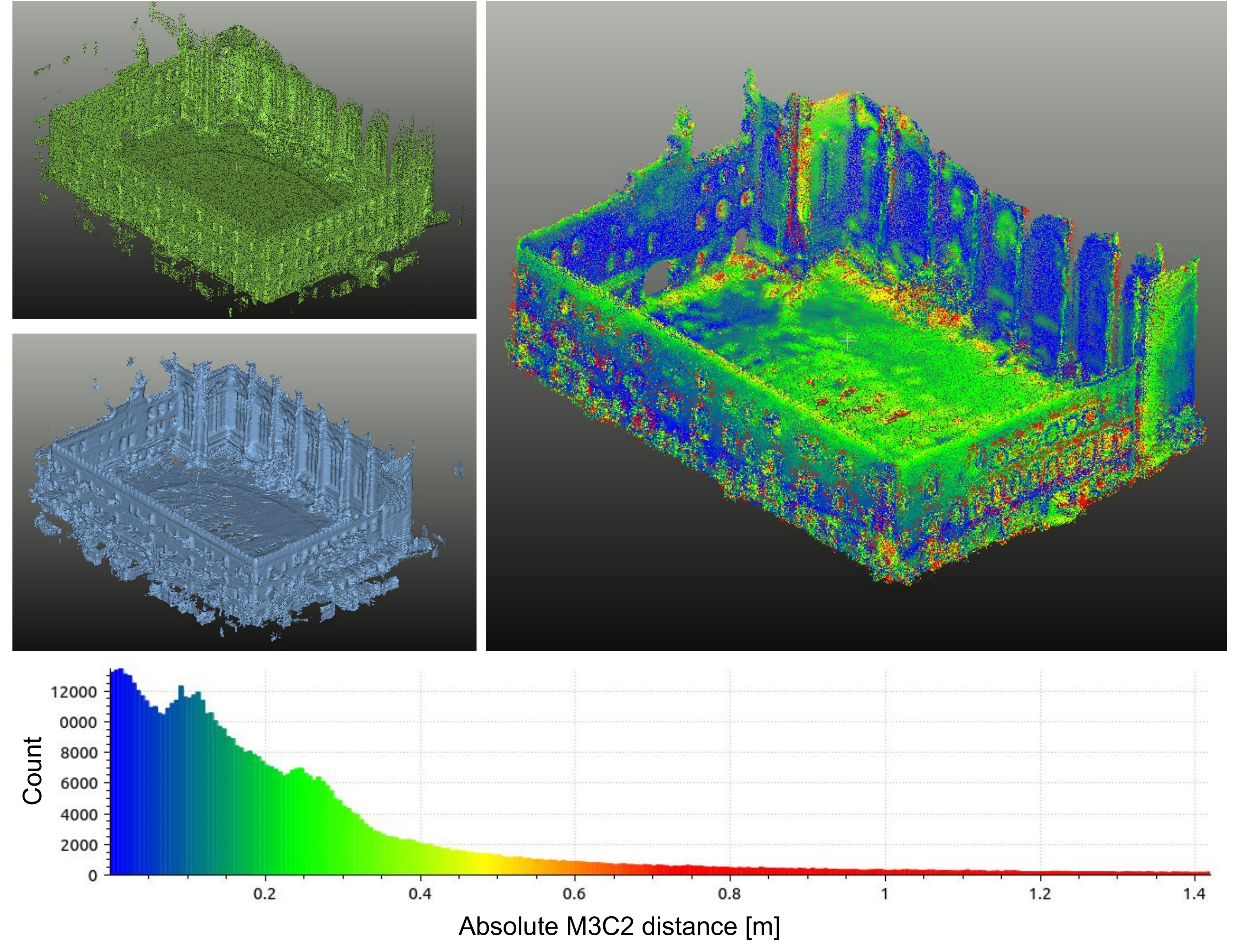}}
\caption{Newer College reconstruction comparison: Top left shows the reference 3D model from the dataset, bottom left shows the output of our architecture and right shows the absolute reconstruction error point cloud with histogram, output of MC3C2 absolute distance computed on the Newer College dataset. 
\\\textbf{Mean = 0.236 m, Std = 0.248 m}}
\label{cloistre_results}
\vspace*{-3mm}
\end{figure}

\begin{figure}[htbp]
\centerline{
\includegraphics[width=0.48\textwidth]{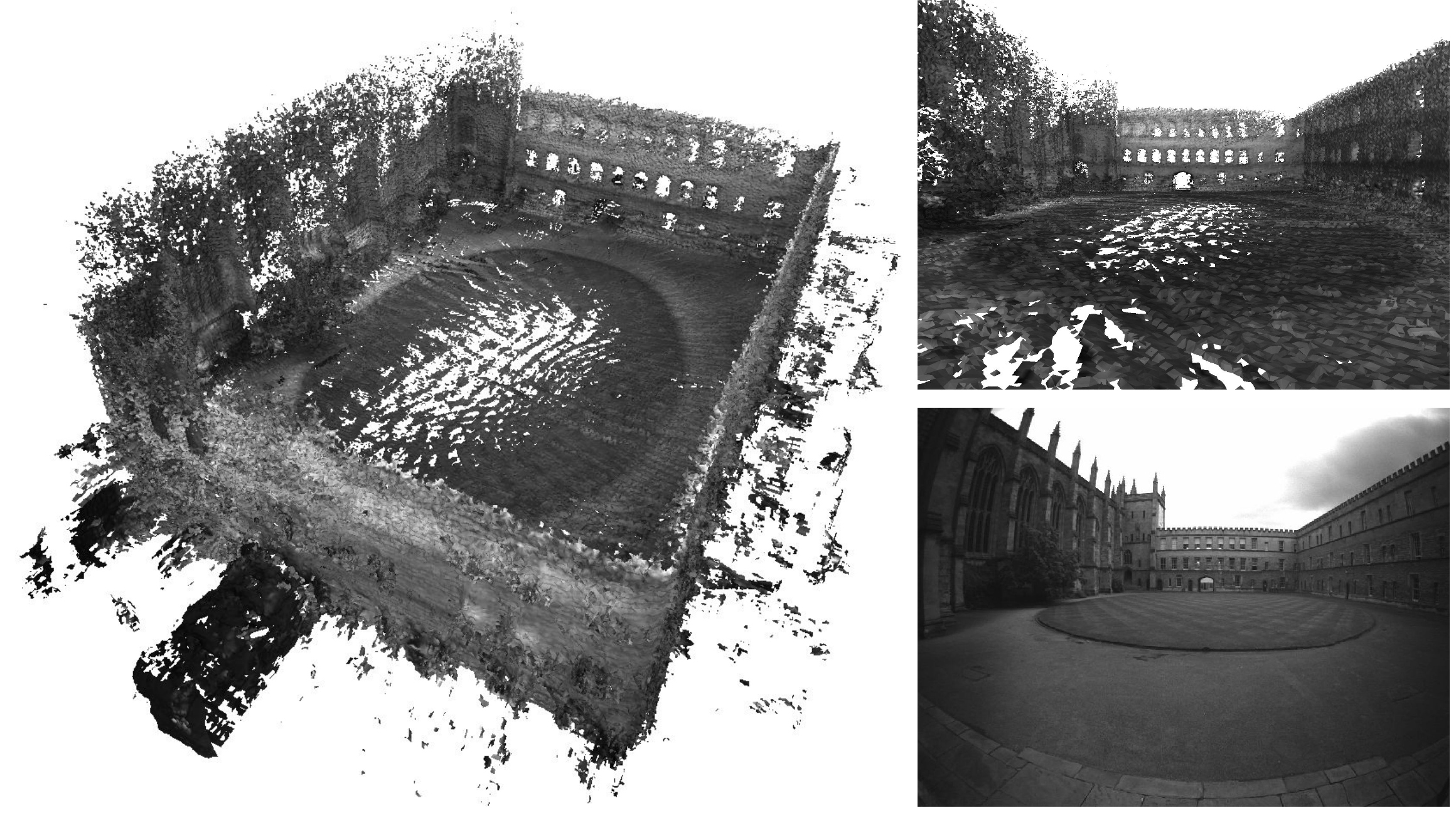}}
\caption{3D mesh with mono camera projection, output of the application of the complete OLCMR architecture on the Newer College Dataset. Left shows the complete reconstructed mesh, right shows a comparison between a camera frame and a corresponding view of the reconstructed mesh.}
\label{newer_full_arch}
\vspace*{-3mm}
\end{figure}

\begin{figure}[htbp]
\centerline{
\includegraphics[width=0.5\textwidth]{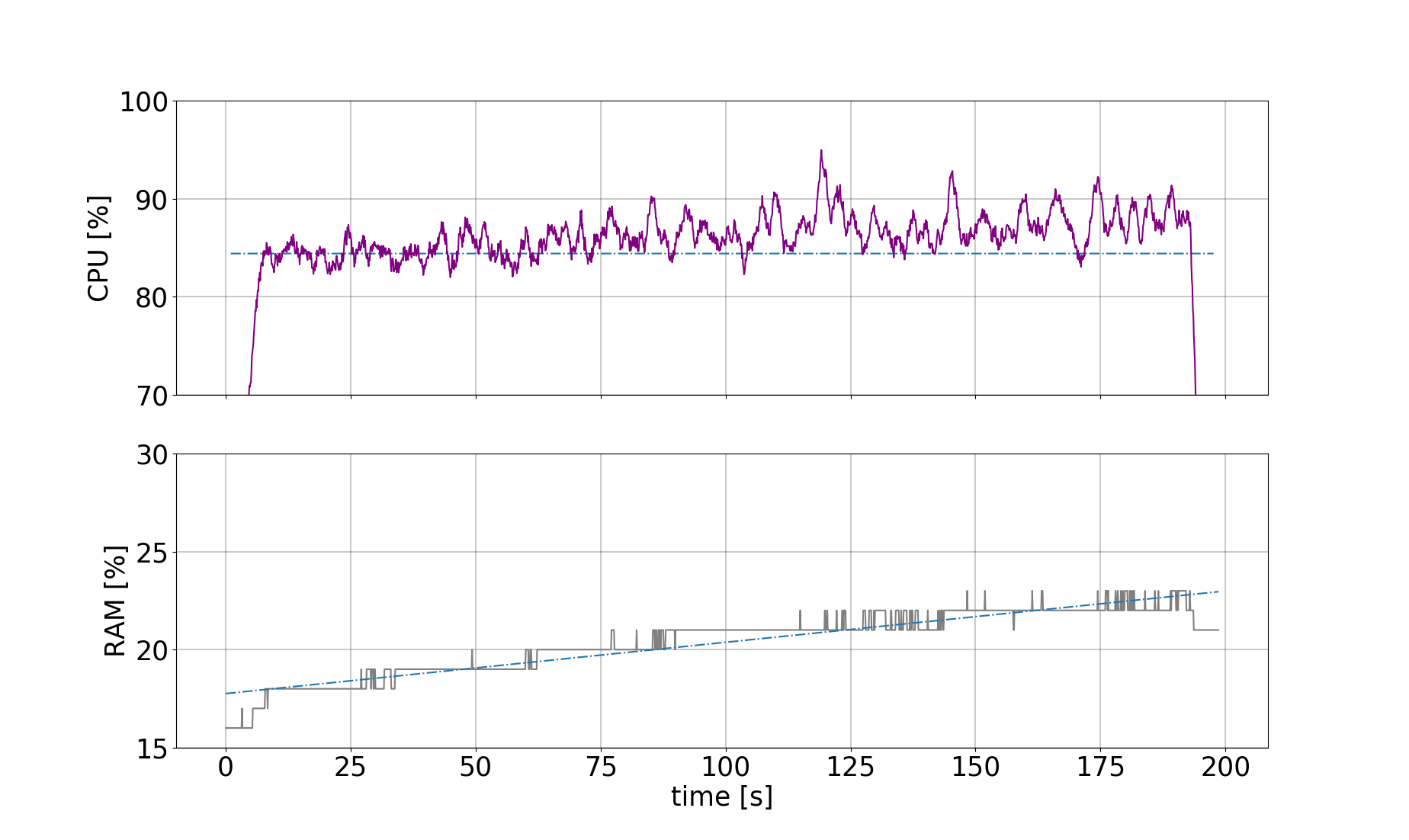}}
\vspace*{-3mm}
\caption{CPU (mean: 84.4 \%, std: 11.7 \%) and RAM (max: 23 \%) usage of OLCMR while running on the Newer College  dataset. Mean CPU usage and linear regression of RAM usage are displayed on relevant figures.
}
\label{newer_cpu}
\vspace*{-5mm}
\end{figure}

The Newer College dataset offers ground-truth 3D meshes of the visited area used to evaluate the quality of our 3D mesh reconstruction. 
After running the mapping pipeline of the architecture (composed of the LiDAR SLAM and Voxblox) on the Newer College dataset, the resulting uncolored mesh is treated and compared to the ground truth model using the CloudCompare open-source software\footnote{ CloudCompare website : \url{https://www.danielgm.net/cc/}}. Both meshes are sampled into dense point clouds which are then manually stripped of aberrant points produced by reflective surfaces (e.g. windows) by applying a planar cut-off beyond said surfaces. The closest point error between the two resulting point clouds is performed using the M3C2 method described in~\cite{M3C2}. Points that have no relevant match are removed. 90\% of our reconstruction model points show a distance error to the ground truth model lesser than 0.54 m. As a comparison, \cite{NC_3D_reconstruction} states that 90\% of the points from their reconstructed model show a distance error lesser than 0.50 m on a similar dataset. The results are illustrated in Figure~\ref{cloistre_results}.
This validates the soundness of the overall LiDAR-based localisation and dense mapping algorithms incorporated in the architecture. The projection of the camera grayscale levels has also been carried out as detailed in Section~\ref{colorisation}, and the associated qualitative result is given in Figure~\ref{newer_full_arch}. The CPU and RAM usage have also been monitored during the processing of the trajectories (see Figure~\ref{newer_cpu}). It turns out that the CPU needs have been successfully adjusted to avoid the saturation of available resources with limited growth over time, and that the RAM requirements increase quite linearly which is a usual behaviour of SLAM systems as the map size increases. By performing a linear fit to the RAM usage, the maximum duration of a mission with similar settings is roughly estimated to 3142 s.  

\section{Evaluation on Field Robots Trajectories}\label{OneraDatasets}

The proposed method obtained good performances on the Newer College dataset, which contains handheld trajectories acquired in a highly-textured environment. To further evaluate the OLCMR architecture in situations closer to its goal applications, we acquired two dedicated experimental datasets produced by tele-operating wheeled ground robots along predefined trajectories and gathering relevant perception data.
These two datasets were respectively produced on-board of a Robotnik Summit-XL robot in a urban environment and an Agilex Scout robot in a countryside environment. These two platforms are four-wheel differentially driven, equipped with on-board CPUs running Ubuntu and ROS, and their respective sensor suites are summarized in Table~\ref{summit_sensors}. OLCMR components main parameters for each dataset are referenced in Table~\ref{params}.
The data processing has been performed on the same computer as for the reference dataset evaluation from Section~\ref{NewerCollege}, which presents similar computational power as the robots' embedded computers.

\begin{table}[htbp]
\caption{Sensors embedded on ONERA-DTIS Summit and Scout robots}
\begin{center}
\begin{tabular}{|c|c|c|}
\hline
\textbf{\textit{Sensor}} & \textbf{\textit{Technical Reference}}& \textbf{\textit{Rate}} \\
\hline
\multicolumn{3}{|c|}{\textbf{Summit}} \\
\hline
Wheel Encoders & Robotnik embedded & 50 Hz \\
\hline
IMU & VectorNav VN-100-t & 400 Hz \\
\hline
LiDAR & Velodyne VLP-16 & 10 Hz \\
\hline
3x RGB Cameras & uEye IDS 3241LE-C & 10 Hz \\
\hline
\multicolumn{3}{|c|}{\textbf{Scout}} \\
\hline
Wheel Encoders & AgileX embedded & 50 Hz \\
\hline
IMU & XSens MTi 300 & 400 Hz \\
\hline
LiDAR & Ouster OS1-32 & 10 Hz \\
\hline
3x RGB Cameras & Basler dart daA1600-60uc & 10 Hz \\
\hline
\end{tabular}
\label{summit_sensors}
\end{center}
\vspace*{-5mm}
\end{table}

\begin{figure}[htbp]
\centerline{\includegraphics[width=0.42\textwidth]{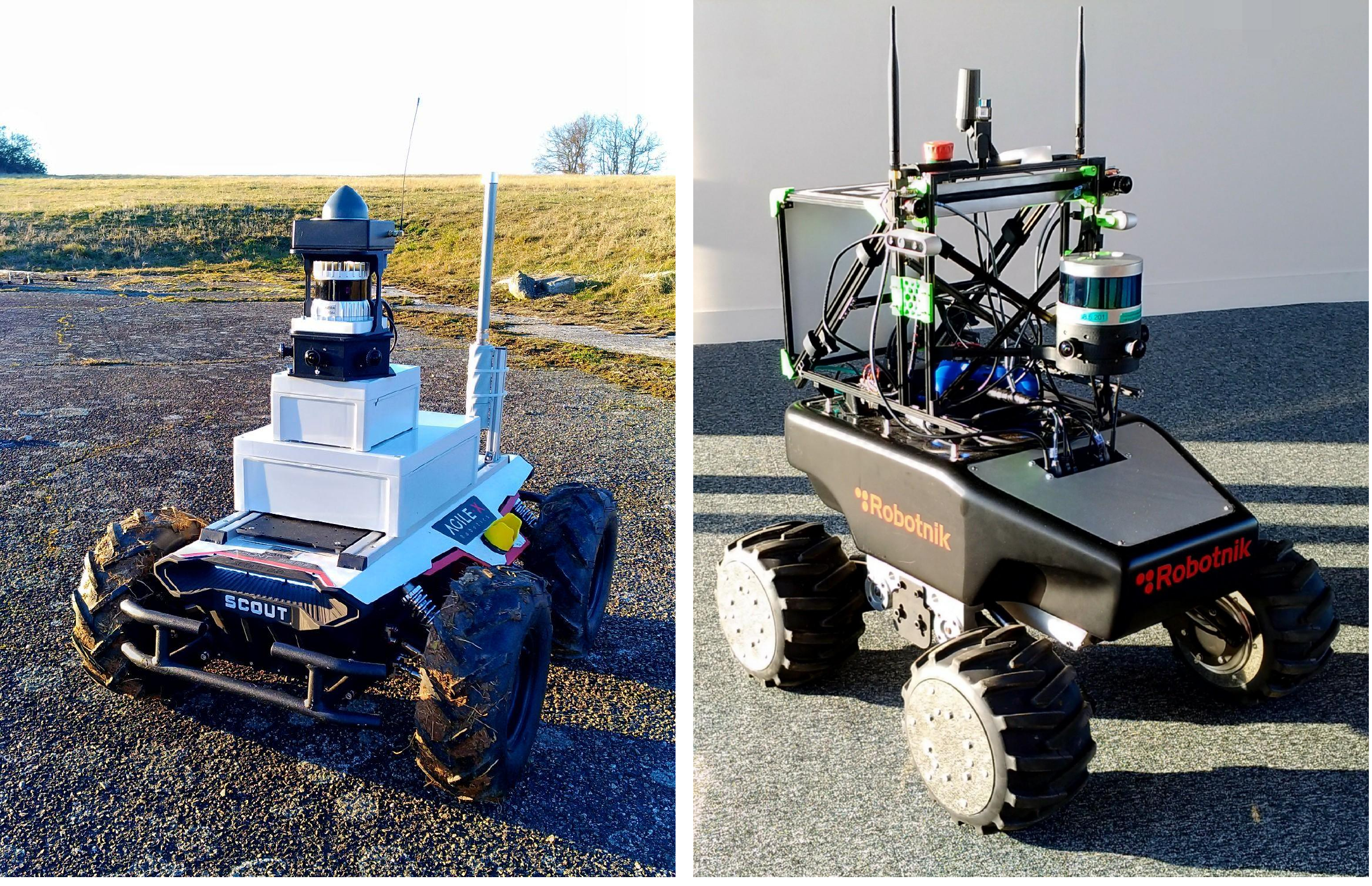}}
\vspace*{-3mm}
\caption{ONERA-DTIS Agilex Scout and Robotnik Summit XL robot setups}
\label{dora_photo}
\vspace*{-5mm}
\end{figure}

\subsection{Localisation Evaluation}
\begin{figure}[htbp]
\vspace*{-3mm}
\centerline{\includegraphics[width=0.47\textwidth]{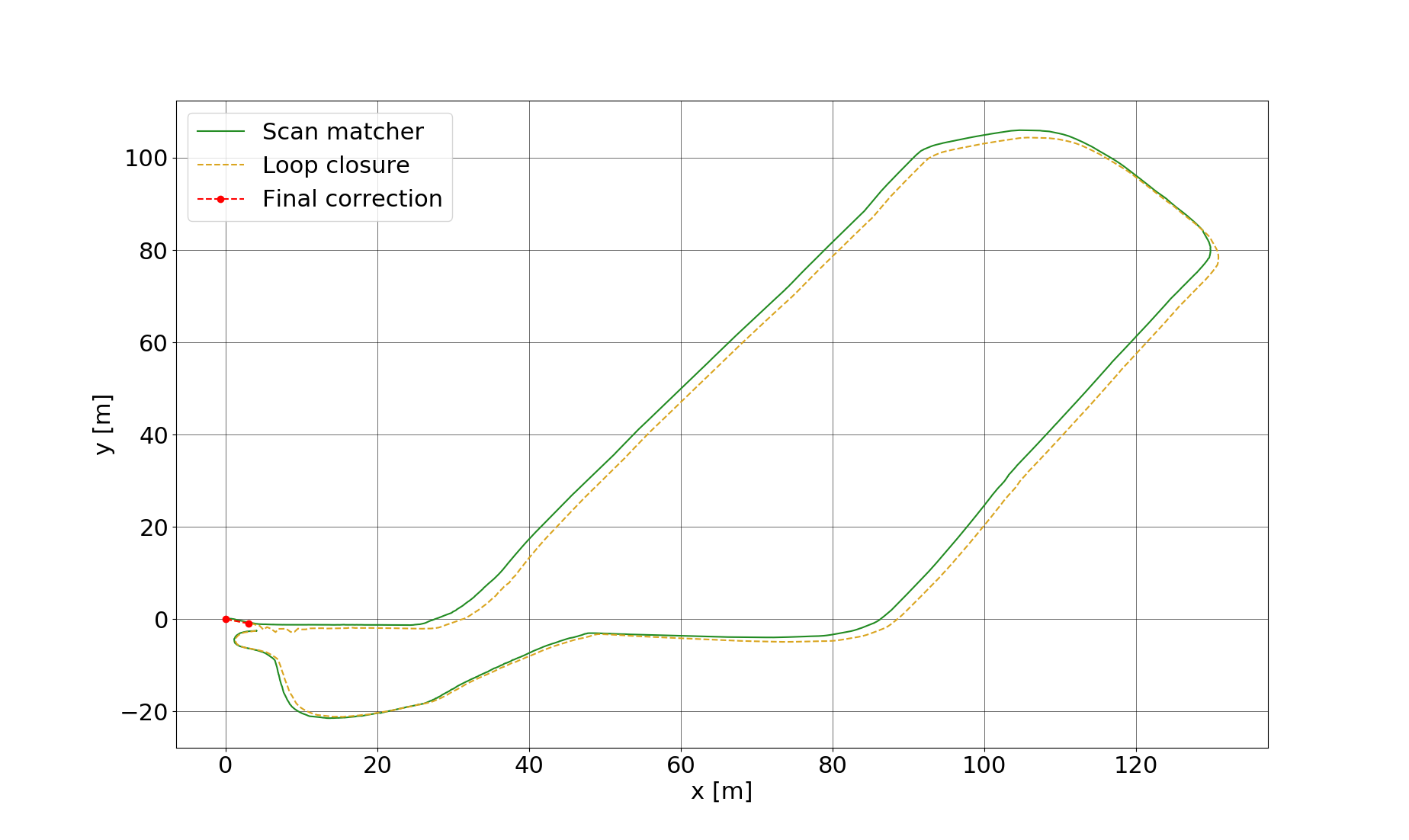}}
\vspace{-0.25cm}
\centerline{
\includegraphics[width=0.47\textwidth]{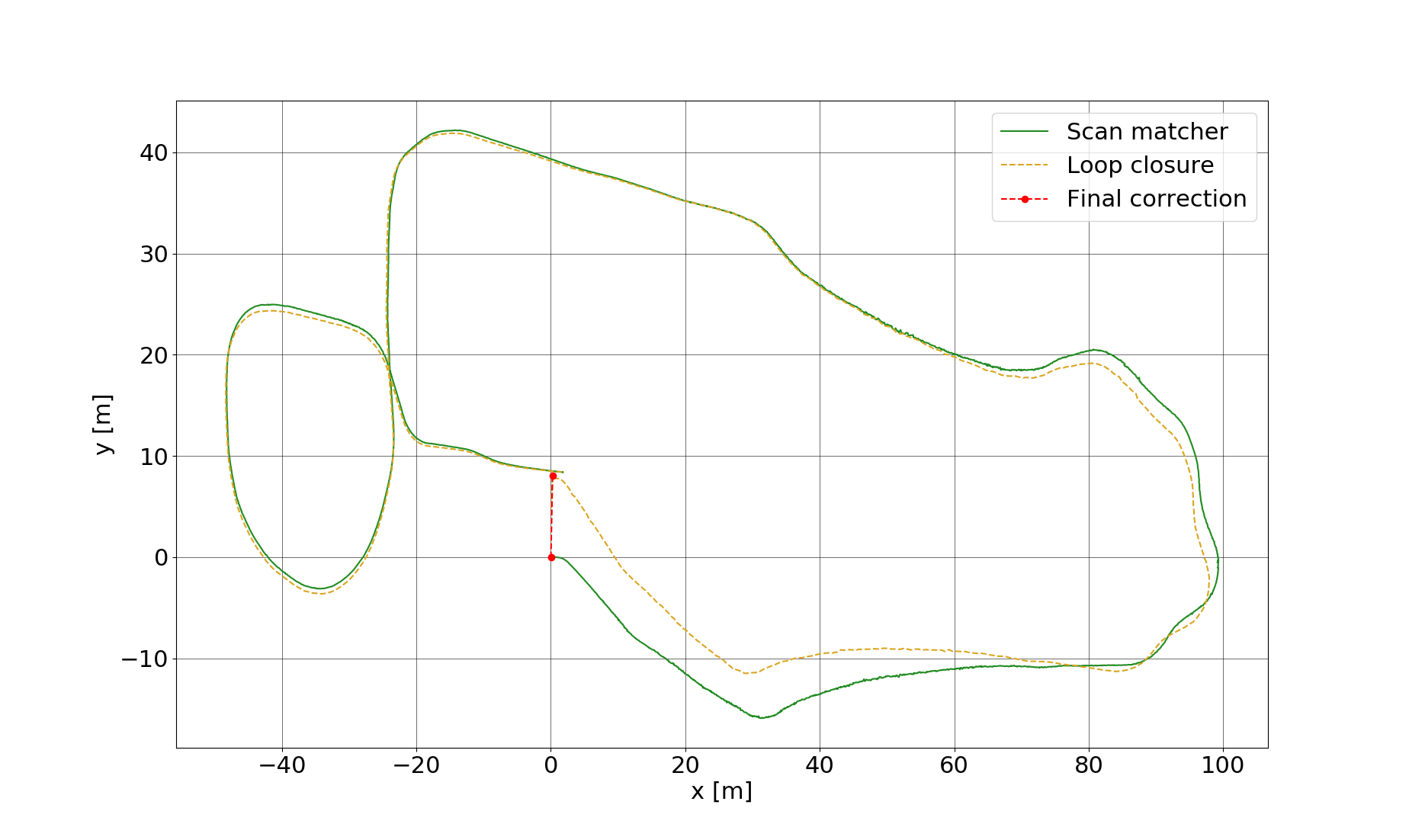}}
\vspace{-0.25cm}
\caption{Robot trajectories on field datasets. Top with the Summit robot, bottom with the Scout robot.}
\label{expe_slam_eval}
\vspace*{-5mm}
\end{figure}

The localisation global performance has been evaluated with respect to the total drift of the trajectories, so this drift is estimated by the difference between the last pose and the first pose computed given the fact that robots were operated in order for the actual ending pose to be approximately equal to the actual starting pose.
The characteristics of the trajectories and the values of the approximated final APE and loop-closure total corrections are summarized in Table~\ref{experimental_slam_compare}, and Figure~\ref{expe_slam_eval} presents the robot estimated trajectories. The overall drift without loop-closure remains between 1 and 2 percents, with a higher value in the less-textured environment. These degradations compared to the previous dataset could be interpreted by the transition from high-quality handheld sensors to robot-mounted lower-grade IMU and LiDAR sensors. These performances remain acceptable to carry out the online 3D dense model rendering process. 

\begin{table}[t]
\caption{SLAM evaluation on field datasets}
\begin{center}
\begin{tabular}{|c|c|}
\hline
\multicolumn{2}{|c|}{\textbf{Summit (urban)}} \\
\hline
Trajectory duration & 516.605 s\\
\hline
Trajectory length & 400.295 m\\
\hline
Scan matcher final APE & 4.794 m\\
\hline
Loop closure final APE & 1.943 m\\
\hline
Final loop closure correction & 3.120 m\\
\hline
\multicolumn{2}{|c|}{\textbf{Scout (countryside)}} \\
\hline
Trajectory duration & 399.399 s\\
\hline
Trajectory length & 407.045 m\\
\hline
Scan matcher final APE & 8.582 m\\
\hline
Loop closure final APE & 1.486 m\\
\hline
Final loop closure correction & 8.034 m\\
\hline
\end{tabular}
\label{experimental_slam_compare}
\end{center}
\vspace{-0.5cm}
\end{table}

\subsection{3D Reconstruction Evaluation}
\begin{figure}[htbp]
\centerline{
\includegraphics[width=0.47\textwidth]{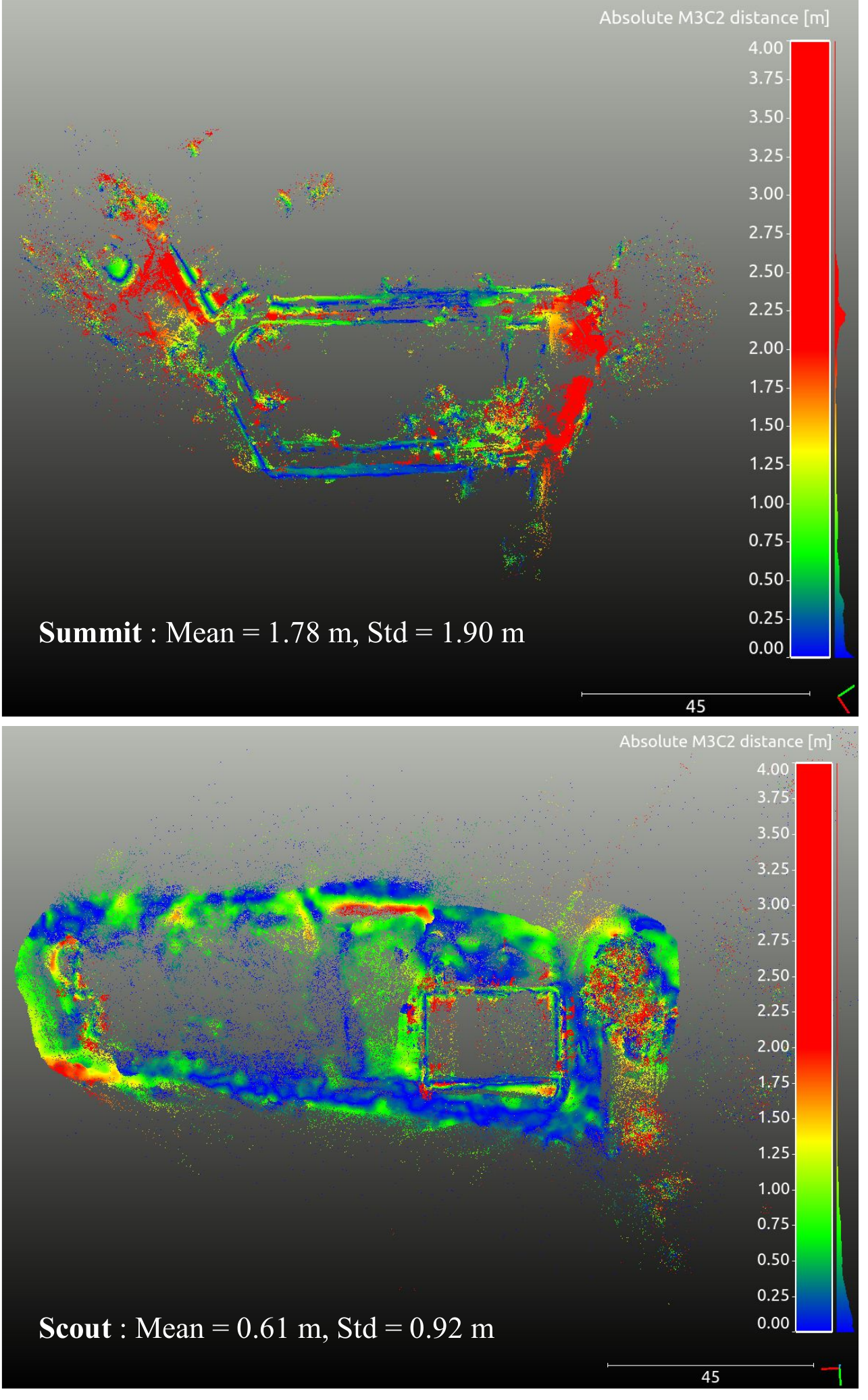}}
\caption{Reconstruction error point cloud map and histogram, output of M3C2 absolute distance computed from Summit (top) and Scout (bottom).}
\label{m3c2_map}
\vspace*{-5mm}
\end{figure}

\begin{figure}[htbp]
\centerline{
\includegraphics[width=0.49\textwidth]{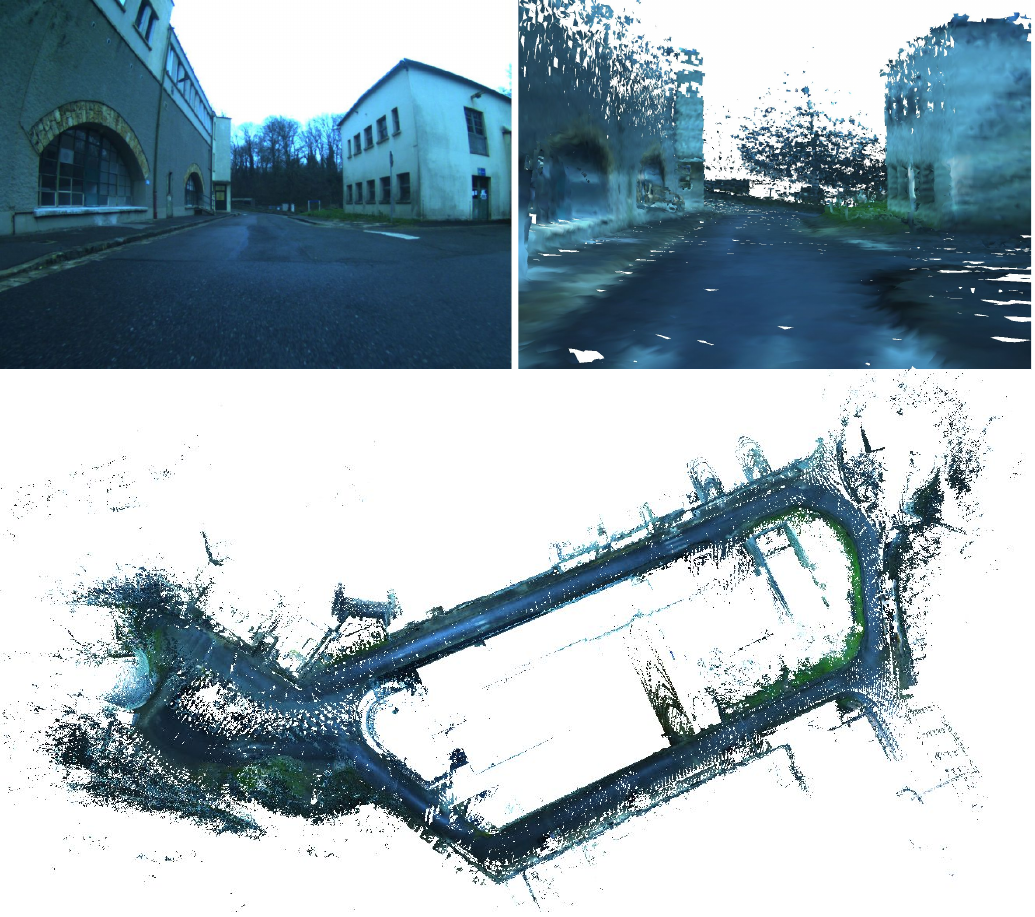}}
\centerline{
\includegraphics[width=0.49\textwidth]{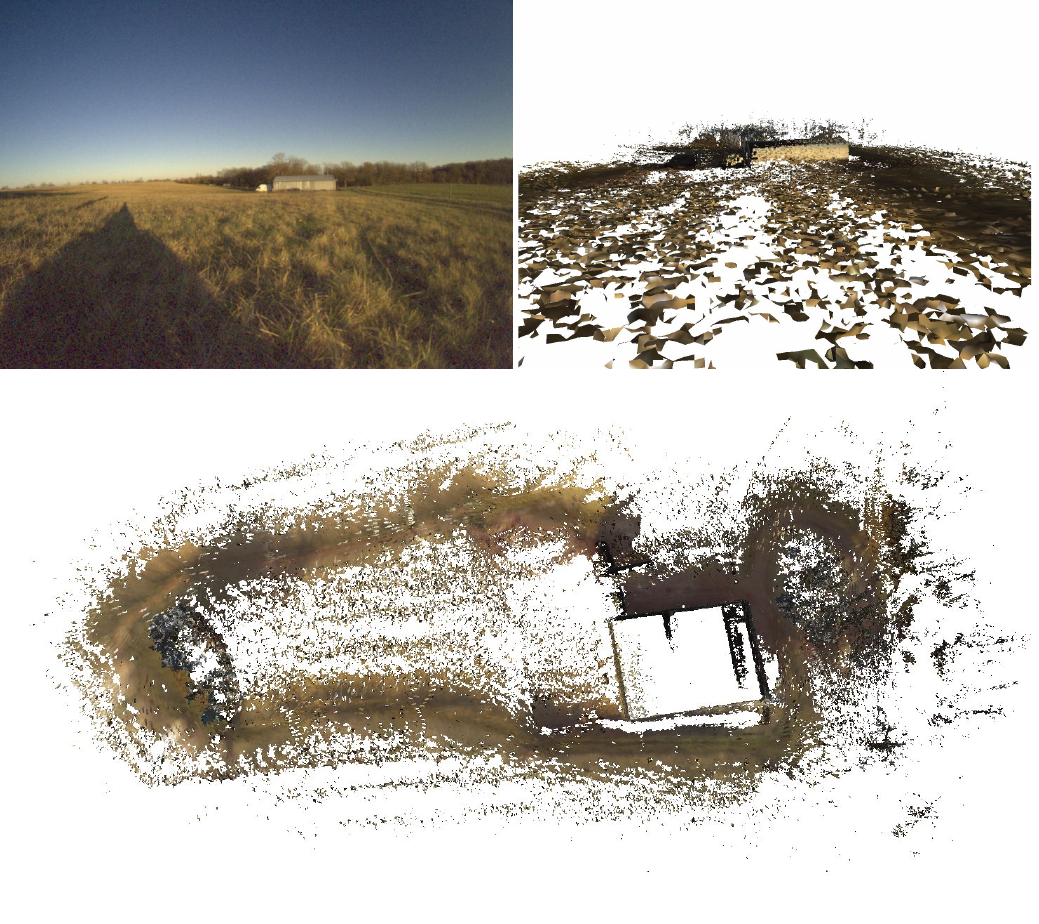}}
\vspace*{-3mm}
\caption{Mesh rendering for Summit robot (urban) dataset (top) and Scout robot (countryside) dataset (bottom).
For each dataset, example of picture with corresponding view of the OLCMR colored mesh and bird-eye view of the global colored mesh are shown.}
\label{render_compare}
\vspace*{-10mm}
\end{figure}

\begin{figure}[htbp]
\centerline{\includegraphics[width=0.5\textwidth]{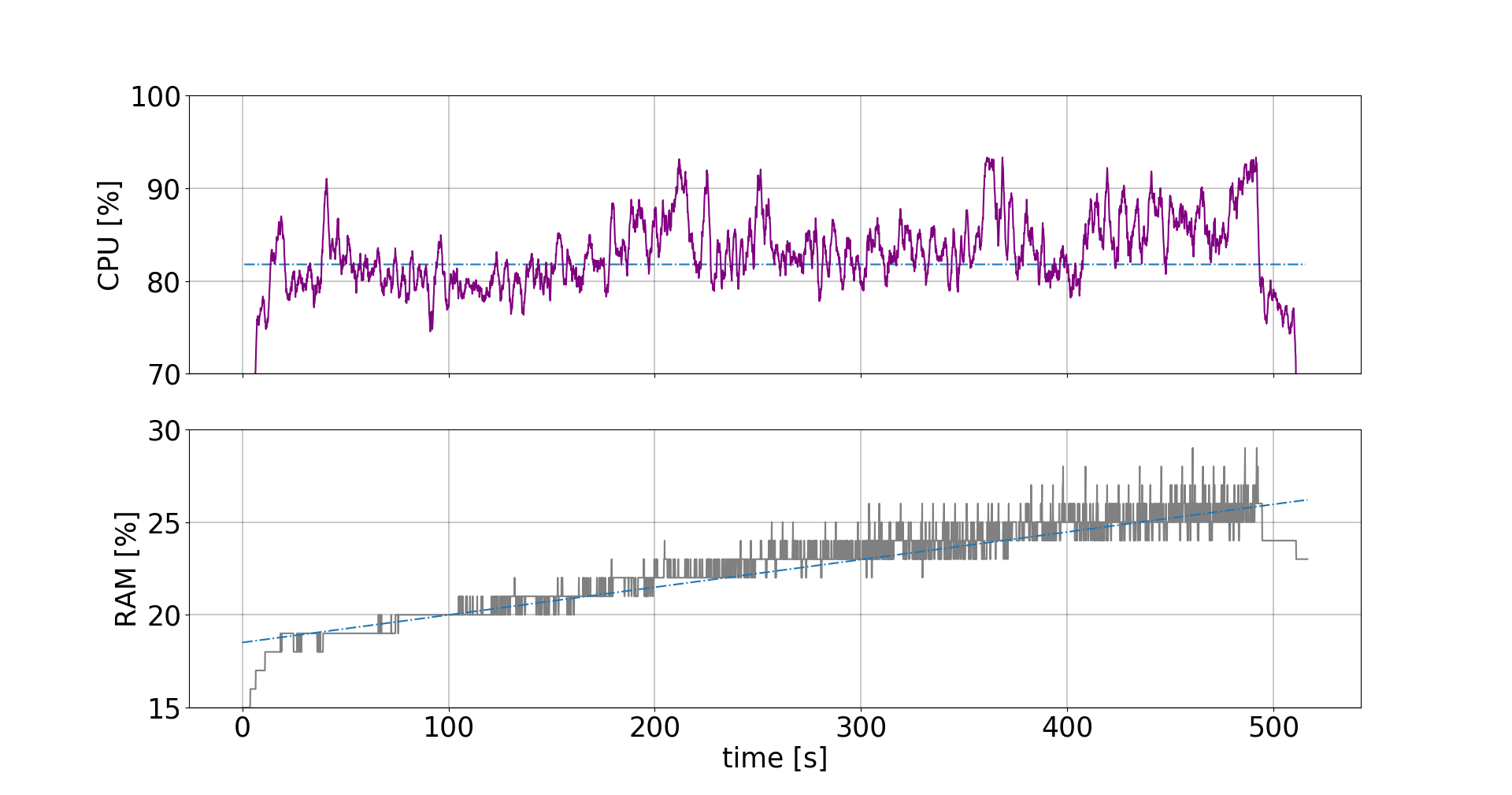}}
\vspace*{-3mm}
\caption{CPU (mean: 81.782 \%, std: 11.7 \%) and RAM (max: 29 \%) usage of OLCMR while on the Summit robot data. Mean CPU usage and linear regression of RAM usage are displayed on relevant figures.}
\vspace*{-3mm}
\label{paslaiseau_cpu}
\end{figure}

Images have been acquired independently from the robot setups in the two new test environments using handheld camera devices, respectively a stereo bench composed of 2 uEye IDS 1241LE-M monocular cameras for the Summit (urban) dataset and a HERO7 GoPro  for the Scout (countryside) dataset. An offline photogrammetric 3D reconstruction has then been obtained using the Structure-From-Motion \textit{Colmap} software~\cite{colmap1}. 
The reconstruction errors between the mapping obtained with the OLCMR architecture using the robots embedded sensors and these Colmap-generated reference models have been analyzed using CloudCompare (see Figure~\ref{m3c2_map}). 
The colored rendering mesh is presented in Figure~\ref{render_compare}. The CPU and RAM usage evaluations (Figure~\ref{paslaiseau_cpu}) show a similar behaviour as on the Newer College dataset, which seem to be compatible with on-board actual deployment for trajectories lengths of one-kilometer order of magnitude.


\begin{table}[htbp]
\caption{OLCMR open-source components main parameters for each experiment}
\begin{center}
\begin{tabular}{|c|c|c|c|}
\hline
\textbf{\textit{parameter}} & \textbf{\textit{Newer College}}& \textbf{\textit{Summit}}& \textbf{\textit{Scout}} \\
\hline
\multicolumn{4}{|c|}{\textbf{LidarSlam scan matcher}} \\
\hline
ndt resolution & 1.5 & 1.0 & 0.8\\
\hline
vg size input & 0.1 & 0.1 & 0.1\\
\hline
vg size map & 0.1 & 0.1 & 0.1\\
\hline
min range & 1.0 & 1.0 & 3.0\\
\hline
max range & 50.0 & 50.0 & 100.0\\
\hline
num targeted cloud & 20 & 20 & 20\\
\hline
\multicolumn{4}{|c|}{\textbf{LidarSlam loop closure}} \\
\hline
ndt resolution & 1.0 & 1.0 & 1.0\\
\hline
voxel leaf size & 0.1 & 0.1 & 0.1\\
\hline
detection period & 7500 & 4000 & 4000\\
\hline
threshold loop closure & 15.0 & 15.0 & 15.0\\
\hline
distance loop closure & 50.0 & 50.0 & 50.0\\
\hline
loop closure search range & 50.0 & 50.0 & 50.0\\
\hline
num submap searched & 10 & 20 & 20\\
\hline
num adj. pose constraints & 10 & 20 & 20\\
\hline
\multicolumn{4}{|c|}{\textbf{Voxblox}} \\
\hline
voxel size & 0.2 & 0.125 & 0.15\\
\hline
voxels per side & 4 & 8 & 8\\
\hline
carving & true & false & false\\
\hline
use free space & false & false & false\\
\hline
method & fast & merged & merged\\
\hline
constant weight & false & false & false\\
\hline
allow clear & false & false & false\\
\hline
min ray length & 2.0 & 0.5 & 0.0\\
\hline
max ray length & 200.0 & 200.0 & 200.0\\
\hline
\end{tabular}
\label{params}
\end{center}
\vspace*{-8mm}
\end{table}


\section{Conclusions and Perspectives}\label{conclusions}

An architecture running in real-time combining LiDAR-based localisation with 3D dense mapping and mesh colourisation using multiple cameras has been proposed in this paper. It is based on recent open-source ROS package implementations of a LiDAR SLAM and a TSDF-based mapping algorithms, with specific developments to assemble the overall pipeline. The full system has been thoroughly evaluated on datasets exhibiting different characteristics, namely the \textit{Newer College} handheld benchmark and two dedicated trajectories acquired with wheeled ground robots in urban and countryside environments. The architecture performed well in all of these conditions, which make it suitable for future field deployment of tele-operated or autonomous robotic exploration.

The loop-closure is currently used solely by the localisation stack.
To improve the development of the OLCMR architecture, it could also be integrated within the mapping stack in a manifold framework such as~\cite{onera_multi_robot_mapping}.
Additional layers could be added to this perception architecture, e.g. for semantic navigation in unknown environments. The next step towards that goal would be to implement semantic segmentation on camera images before projecting them into the 3D space in order to create semantic maps that could be used by the robot for safer and better autonomous navigation.

\bibliography{bibliographie}

\begin{thebibliography}{10}

\bibitem{lluvia2021active}
Iker Lluvia, Elena Lazkano, and Ander Ansuategi.
\newblock Active mapping and robot exploration: A survey.
\newblock {\em Sensors}, 21(7):2445, 2021.

\bibitem{fuentes2015visual}
Jorge Fuentes-Pacheco, Jos{\'e} Ruiz-Ascencio, and Juan~Manuel
  Rend{\'o}n-Mancha.
\newblock Visual simultaneous localization and mapping: a survey.
\newblock {\em Artificial intelligence review}, 43(1):55--81, 2015.

\bibitem{yasuda2020autonomous}
Yuri~DV Yasuda, Luiz Eduardo~G Martins, and Fabio~AM Cappabianco.
\newblock Autonomous visual navigation for mobile robots: A systematic
  literature review.
\newblock {\em ACM Computing Surveys (CSUR)}, 53(1):1--34, 2020.

\bibitem{merzlyakov2021comparison}
Alexey Merzlyakov and Steve Macenski.
\newblock A comparison of modern general-purpose visual {SLAM} approaches.
\newblock In {\em IEEE/RSJ International Conference on Intelligent Robots and
  Systems (IROS)}, pages 9190--9197, 2021.

\bibitem{elhousni2020survey}
Mahdi Elhousni and Xinming Huang.
\newblock A survey on {3D LiDAR} localization for autonomous vehicles.
\newblock In {\em IEEE Intelligent Vehicles Symposium (IV)}, pages 1879--1884,
  2020.

\bibitem{zou2021comparative}
Qin Zou, Qin Sun, Long Chen, Bu~Nie, and Qingquan Li.
\newblock A comparative analysis of {LiDAR SLAM}-based indoor navigation for
  autonomous vehicles.
\newblock {\em IEEE Transactions on Intelligent Transportation Systems}, 2021.

\bibitem{rtab-map}
Mathieu Labbé and François Michaud.
\newblock {RTAB-Map} as an open-source lidar and visual simultaneous
  localization and mapping library for large-scale and long-term online
  operation.
\newblock {\em Journal of Field Robotics}, 36(2):416--446, 2019.

\bibitem{NC_3D_reconstruction}
Yiduo Wang, Nils Funk, Milad Ramezani, Sotiris Papatheodorou, Marija Popović,
  Marco Camurri, Stefan Leutenegger, and Maurice Fallon.
\newblock Elastic and efficient {LiDAR} reconstruction for large-scale
  exploration tasks.
\newblock In {\em IEEE International Conference on Robotics and Automation
  (ICRA)}, pages 5035--5041, 2021.

\bibitem{debeunne2020review}
C{\'e}sar Debeunne and Damien Vivet.
\newblock A review of visual-lidar fusion based simultaneous localization and
  mapping.
\newblock {\em Sensors}, 20(7):2068, 2020.

\bibitem{kim2018slam}
Pileun Kim, Jingdao Chen, and Yong~K Cho.
\newblock {SLAM-driven} robotic mapping and registration of {3D} point clouds.
\newblock {\em Automation in Construction}, 89:38--48, 2018.

\bibitem{waechter2014let}
Michael Waechter, Nils Moehrle, and Michael Goesele.
\newblock Let there be color! large-scale texturing of {3D} reconstructions.
\newblock In {\em European Conference on Computer Vision (ECCV)}, pages
  836--850, 2014.

\bibitem{chan2021post}
Ting~On Chan, Hang Xiao, Lixin Liu, Yeran Sun, Tingting Chen, Wei Lang, and
  Ming~Ho Li.
\newblock A post-scan point cloud colorization method for cultural heritage
  documentation.
\newblock {\em ISPRS International Journal of Geo-Information}, 10(11):737,
  2021.

\bibitem{aguiar2020localization}
Andr{\'e}~Silva Aguiar, Filipe~Neves dos Santos, Jos{\'e}~Boaventura Cunha,
  H{\'e}ber Sobreira, and Armando~Jorge Sousa.
\newblock Localization and mapping for robots in agriculture and forestry: A
  survey.
\newblock {\em Robotics}, 9(4):97, 2020.

\bibitem{early_online_mesh_construction}
Marc Pollefeys and al.
\newblock Detailed real-time urban {3D} reconstruction from video.
\newblock {\em International Journal of Computer Vision}, 78:143--167, 07 2008.

\bibitem{chisel}
M.~Klingensmith, I.~Dryanovski, S.~Srinivasa, and J.~Xiao.
\newblock {CHISEL: Real Time Large Scale 3D Reconstruction Onboard a Mobile
  Device using Spatially Hashed Signed Distance Fields}.
\newblock In {\em Robotics: Science and Systems XI}, volume~4, 2015.

\bibitem{SDF}
Brian Curless and Marc Levoy.
\newblock A volumetric method for building complex models from range images.
\newblock {\em Proceedings of the 23rd annual conference on Computer graphics
  and interactive techniques}, 1996.

\bibitem{hashing}
M.~Nie{\ss}ner, M.~Zollh\"ofer, S.~Izadi, and M.~Stamminger.
\newblock Real-time {3D} reconstruction at scale using voxel hashing.
\newblock {\em ACM Transactions on Graphics (TOG)}, 2013.

\bibitem{real_time_3D_mesh_reconstruction}
Enrico Piazza, Andrea Romanoni, and Matteo Matteucci.
\newblock Real-time {CPU}-based large-scale {3D} mesh reconstruction.
\newblock {\em arXiv preprint arXiv:1801.05230}, 2018.

\bibitem{kitti}
Andreas Geiger, Philip Lenz, and Raquel Urtasun.
\newblock Are we ready for autonomous driving? the {KITTI} vision benchmark
  suite.
\newblock In {\em IEEE Conference on Computer Vision and Pattern Recognition},
  pages 3354--3361, 2012.

\bibitem{surfel_mapping}
Kaixuan Wang, Fei Gao, and Shaojie Shen.
\newblock Real-time scalable dense surfel mapping.

\bibitem{urban-mapping}
Andrea Romanoni, Daniele Fiorenti, and Matteo Matteucci.
\newblock Mesh-based {3D} textured urban mapping.
\newblock In {\em IEEE/RSJ International Conference on Intelligent Robots and
  Systems (IROS)}, pages 3460--3466, 2017.

\bibitem{mesh-reconstruction}
Mohamed Boussaha, Bruno Vallet, and Patrick Rives.
\newblock Large scale textured mesh reconstruction from mobile mapping images
  and {LiDAR} scans.
\newblock In {\em International Society for Photogrammetry and Remote Sensing},
  pages 49--56, 2018.

\bibitem{onera_multi_robot_mapping}
Thibaud Duhautbout, Julien Moras, and Julien Marzat.
\newblock Distributed {3D TSDF} manifold mapping for multi-robot systems.
\newblock In {\em European Conference on Mobile Robots (ECMR)}, 2019.

\bibitem{ScalableFusion}
Simon Schreiberhuber, Johann Prankl, Timothy Patten, and Markus Vincze.
\newblock Scalablefusion: High-resolution mesh-based real-time {3D}
  reconstruction.
\newblock In {\em International Conference on Robotics and Automation (ICRA)},
  pages 140--146, 2019.

\bibitem{newer_college}
Milad Ramezani, Yiduo Wang, Marco Camurri, David Wisth, Matias Mattamala, and
  Maurice Fallon.
\newblock The newer college dataset: Handheld {LiDAR}, inertial and vision with
  ground truth.
\newblock {\em IEEE/RSJ International Conference on Intelligent Robots and
  Systems (IROS)}, 2020.

\bibitem{newer_college_2}
Lintong Zhang, Marco Camurri, and Maurice Fallon.
\newblock Multi-camera {LiDAR} inertial extension to the {Newer College}
  dataset.
\newblock {\em arXiv preprint arXiv:2112.08854}, 2021.

\bibitem{kalibr}
J{\'e}r{\^o}me Maye, Paul Furgale, and Roland Siegwart.
\newblock Self-supervised calibration for robotic systems.
\newblock In {\em IEEE Intelligent Vehicles Symposium (IV)}, pages 473--480,
  2013.

\bibitem{LidarSlam}
Lidarslam ros2.
\newblock \url{"https://github.com/rsasaki0109/lidarslam_ros2"}, 2020.

\bibitem{voxblox}
Helen Oleynikova, Zachary Taylor, Marius Fehr, Roland Siegwart, and Juan Nieto.
\newblock Voxblox: Incremental {3D} euclidean signed distance fields for
  on-board {MAV} planning.
\newblock In {\em IEEE/RSJ International Conference on Intelligent Robots and
  Systems (IROS)}, 2017.

\bibitem{stavrinos2021ros2}
George Stavrinos.
\newblock {ROS2} for {ROS1} users.
\newblock In {\em Robot Operating System (ROS)}, pages 31--42. Springer, 2021.

\bibitem{liosam}
Tixiao Shan, Brendan Englot, Drew Meyers, Wei Wang, Carlo Ratti, and Rus
  Daniela.
\newblock {LIO-SAM}: Tightly-coupled lidar inertial odometry via smoothing and
  mapping.
\newblock In {\em IEEE/RSJ International Conference on Intelligent Robots and
  Systems (IROS)}, pages 5135--5142. IEEE, 2020.

\bibitem{recent_icp}
Juyong Zhang, Yuxin Yao, and Bailin Deng.
\newblock Fast and robust iterative closest point.
\newblock {\em IEEE Transactions on Pattern Analysis and Machine Intelligence},
  2021.

\bibitem{legoloam}
Tixiao Shan and Brendan Englot.
\newblock {LeGO-LOAM}: Lightweight and ground-optimized {LiDAR} odometry and
  mapping on variable terrain.
\newblock In {\em IEEE/RSJ International Conference on Intelligent Robots and
  Systems (IROS)}, pages 4758--4765. IEEE, 2018.

\bibitem{ndt}
Peter Biber and Wolfgang Stra{\ss}er.
\newblock The normal distributions transform: A new approach to laser scan
  matching.
\newblock In {\em IEEE/RSJ International Conference on Intelligent Robots and
  Systems (IROS)}, volume~3, pages 2743--2748, 2003.

\bibitem{g2o}
Rainer Kümmerle, Giorgio Grisetti, Hauke Strasdat, Kurt Konolige, and Wolfram
  Burgard.
\newblock G$^{2}$o: A general framework for graph optimization.
\newblock In {\em 2011 IEEE International Conference on Robotics and
  Automation}, pages 3607--3613, 2011.

\bibitem{octomap}
Armin Hornung, Kai~M. Wurm, Maren Bennewitz, Cyrill Stachniss, and Wolfram
  Burgard.
\newblock {OctoMap}: An efficient probabilistic {3D} mapping framework based on
  octrees.
\newblock {\em Autonomous Robots}, 2013.

\bibitem{projection}
Robert~M Haralick.
\newblock Using perspective transformations in scene analysis.
\newblock {\em Computer Graphics and Image Processing}, 13(3):191--221, 1980.

\bibitem{marulan_dataset}
Thierry Peynot, Steve Scheding, and Sami Terho.
\newblock The {Marulan} data sets: Multi-sensor perception in a natural
  environment with challenging conditions.
\newblock {\em The International Journal of Robotics Research},
  29(13):1602--1607, 2010.

\bibitem{husky_dataset}
Abdulbaki Aybakan, Garen Haddeler, M.~Caner Akay, Osman Ervan, and Hakan
  Temeltas.
\newblock A {3D LiDAR} dataset of {ITU} heterogeneous robot team.
\newblock In {\em Proceedings of the 2019 5th International Conference on
  Robotics and Artificial Intelligence}, ICRAI '19, page 12–17, 2019.

\bibitem{slam_evaluation}
Jürgen Sturm, Nikolas Engelhard, Felix Endres, Wolfram Burgard, and Daniel
  Cremers.
\newblock A benchmark for the evaluation of {RGB-D SLAM} systems.
\newblock In {\em IEEE/RSJ International Conference on Intelligent Robots and
  Systems (IROS)}, pages 573--580, 2012.

\bibitem{nc_slam_compare}
Lintong Zhang, David Wisth, Marco Camurri, and Maurice Fallon.
\newblock Balancing the budget: Feature selection and tracking for multi-camera
  visual-inertial odometry.
\newblock {\em IEEE Robotics and Automation Letters}, 7(2):1182–1189, 2022.

\bibitem{quenzel2021realtime}
Jan Quenzel and Sven Behnke.
\newblock Real-time multi-adaptive-resolution-surfel 6d lidar odometry using
  continuous-time trajectory optimization, 2021.

\bibitem{open_vins}
Patrick Geneva, Kevin Eckenhoff, Woosik Lee, Yulin Yang, and Guoquan Huang.
\newblock {OpenVINS}: A research platform for visual-inertial estimation.
\newblock In {\em IEEE International Conference on Robotics and Automation
  (ICRA)}, Paris, France, 2020.

\bibitem{LOAM}
Ji~Zhang and Sanjiv Singh.
\newblock Loam : Lidar odometry and mapping in real-time.
\newblock {\em Robotics: Science and Systems Conference (RSS)}, pages 109--111,
  01 2014.

\bibitem{orb_slam3}
Carlos Campos, Richard Elvira, Juan J.~Gomez Rodriguez, Jose~M.M Montiel, and
  Juan~D. Tardos.
\newblock {ORB-SLAM3}: An accurate open-source library for visual,
  visual–inertial, and multimap {SLAM}.
\newblock {\em IEEE Transactions on Robotics}, 37(6):1874–1890, 2021.

\bibitem{F-LOAM}
Han Wang, Chen Wang, Chun-Lin Chen, and Lihua Xie.
\newblock F-loam: Fast lidar odometry and mapping.
\newblock {\em 2021 IEEE/RSJ International Conference on Intelligent Robots and
  Systems (IROS)}, Sep 2021.

\bibitem{SuMa}
Jens Behley and Cyrill Stachniss.
\newblock Efficient surfel-based slam using 3d laser range data in urban
  environments.
\newblock 06 2018.

\bibitem{M3C2}
Mike~R. James, Stuart Robson, and Mark~W. Smith.
\newblock {3-D} uncertainty-based topographic change detection with
  structure-from-motion photogrammetry: precision maps for ground control and
  directly georeferenced surveys.
\newblock {\em Earth Surface Processes and Landforms}, 42(12):1769--1788, 2017.

\bibitem{colmap1}
Johannes~Lutz Sch\"{o}nberger and Jan-Michael Frahm.
\newblock Structure-from-motion revisited.
\newblock In {\em Conference on Computer Vision and Pattern Recognition
  (CVPR)}, 2016.

\end{thebibliography}
\bibliographystyle{unsrt}

\end{document}